\documentclass[runningheads]{llncs}
\usepackage[T1]{fontenc}
\usepackage{graphicx}
\usepackage{amsmath}
\usepackage{amssymb}
\usepackage{hyperref}
\usepackage{listings}
\usepackage{framed}
\usepackage{float}
\usepackage{subfig}
\usepackage{anyfontsize}
\usepackage{caption} 
\usepackage{xcolor}%
\begin{document}
\title{SCALOFT: An Initial Approach for Situation Coverage-Based Safety Analysis of an Autonomous Aerial Drone in a Mine Environment}
\titlerunning{Situation Coverage based Safety Testing of AAD}
%
\author{Nawshin Mannan Proma\inst{1}\orcidID{0000-0002-8869-3977} \and
Victoria J Hodge\inst{1}\orcidID{0000-0002-2469-0224} \and
Rob Alexander\inst{1}\orcidID{0000-0003-3818-0310}}
\authorrunning{N.M. Proma et al.}
%
\institute{University of York, Heslington, United Kingdom
\email{nawshinmannan.proma@york.ac.uk, victoria.hodge@york.ac.uk, rob.alexander@york.ac.uk}\\
}
\maketitle

\begin{abstract}

The safety of autonomous systems in dynamic and hazardous environments poses significant challenges. This paper presents a testing approach named SCALOFT for systematically assessing the safety of an autonomous aerial drone in a mine. SCALOFT provides a framework for developing diverse test cases, real-time monitoring of system behaviour, and detection of safety violations. Detected violations are then logged with unique identifiers for detailed analysis and future improvement. SCALOFT helps build a safety argument by monitoring situation coverage and calculating a final coverage measure. We have evaluated the performance of this approach by deliberately introducing seeded faults into the system and assessing whether SCALOFT is able to detect those faults. For a small set of plausible faults, we show that SCALOFT is successful in this.

\end{abstract}

\section{Introduction}

Autonomous aerial drones (AAD) have gained popularity due to their ability to operate autonomously with minimal or no human intervention thus minimising human risk and offering cost-effective solutions \cite{shakhatreh2019unmanned}. Their applications include search and rescue missions, building inspections and navigating challenging environments such as underground mines. However, the dynamic nature of these operational environments necessitates the development of a systematic testing approach to ensure the safety of both the drones and their operating contexts.

Establishing a detailed safety assessment process during the design phase~\cite{hodge2021deep} of AAD applications is difficult, especially where humans will be present, such as in underground mines. Such a process would need to identify a representative set of potential failure situations during AAD operations, assess their consequences, and define mitigation measures to minimize risks. For example, even if a drone can fly safely in different light conditions, we still need to make sure it follows  the safety requirements when it is flying in complicated situations and ensure the risk posed is as low as reasonably practicable. 

In traditional software testing, it is common to use coverage measures to check if all parts of the software were tested. This could be as simple as making sure every line of code was run at least once. However, these measures have been criticised for not capturing all the important aspects of the software \cite{Rob2015}. They might miss things like external factors that can affect how the software behaves. 

Traditional safety testing employs a variety of coverage techniques, such as system coverage, requirement coverage, and scenario coverage (refer to Table \ref{tab1} for an AAD example of how these differ from one another). But testing autonomous software is always tricky because AADs can encounter all sorts of different situations, such as obstacles or interactions with people, in a constantly changing environment. To test autonomous system  software in both expected and unexpected situations, it is necessary to consider a sufficient range of situations while testing \cite{Rob2015,Hawkins2019}.
\begin{table}[h!]
\fontsize{8pt}{8pt}\selectfont
\centering
\caption{Coverage based safety testing}\label{tab1}
\renewcommand{\arraystretch}{1.5}
\setlength{\tabcolsep}{1.8pt} 
\begin{tabular}{p{3cm}p{2.6cm}p{2.6cm}p{3.7cm}}
\hline
Testing Approach & Focus & Key Attributes & Example of AAD Testing in Mine \\
\hline
System Coverage   \cite{pei2017deepxplore},\cite{tian2018deeptest}, \cite{katz2017reluplex},\cite{huang2017safety},\cite{kurakin2018adversarial} & Thorough testing of system components and interactions & Test all components and their interactions & Ensuring sensors, decision-making algorithms, and other systems work together effectively to avoid collisions \\ 
Requirement Coverage \cite{Rob2015} & Verifying compliance with specified safety requirements & Focuses on fulfilling explicit requirements & Ensuring the requirement that AAD slows within a specified distance when a person is detected inside the mine \\ 
Scenario Coverage \cite{ulbrich2015defining}, \cite{abdessalem2018testing}, \cite{iqbal2015environment}, \cite{micskei2012concept}, \cite{nguyen2012evolutionary} & Performance in predefined real-world scenarios & Evaluates system behavior in a sequence of events & Testing drone’s behavior at predefined waypoints in the mine \\ 
Situation Coverage \cite{Rob2015}, \cite{Tahir2022}, \cite{nawshin2023}, \cite{proma2024situation}, \cite{majzik2019towards},\cite{babikian2020} & System’s adaptability in dynamic, real-time conditions & Assess system’s ability to handle unexpected situations & Evaluating how the drone responds to random obstacles inside the mine \\
\hline
\end{tabular}
\end{table}Situation coverage-based safety testing \cite{Rob2015} assesses the system's performance in dynamic, real-time situations, testing its robustness and adaptability to new situations. While this approach effectively tests the system's ability to handle unforeseen events, it is impossible to ensure exhaustive coverage of all possible situations in system-level testing. Some recent literature describes how to find representative situations for autonomous vehicles (AVs) using a situation-coverage-based approach \cite{nawshin2023,Tahir2022}. Currently, the methods can only generate simple test situations. These might help find bugs, but they do not cover all possible situations that can occur in the real world. Also, while some research has been conducted on situation coverage testing for AVs \cite{proma2024situation,ryan2024safety}, there is a lack of comprehensive studies addressing situation generation and safety testing of AADs to address the unique challenges of underground environments. This gap highlights the need for dedicated research work to develop and validate a situation coverage-based safety testing approach that addresses the operational requirements of AADs in underground environments such as mines. 

Due to the identified research gaps, this study aims to explore the construction of a situation hyperspace — a conceptual model that systematically organizes various factors, including environmental conditions, system states, and external influences, to comprehensively represent the possible combinations of operating scenarios for AAD based on the Operational Domain Model (ODM). ODM defines the specific conditions under which an autonomous system is designed to function safely to establish the system’s operational boundaries (discussed next in section \ref{sec:ODM}). Additionally, the research will assess whether the test cases derived from this situation hyperspace provide adequate coverage of the potential situations an AAD system may encounter. Before conducting situation coverage-based testing, this study assumes that component-level coverage has already been achieved and is sufficient. While ensuring component-level coverage is a significant challenge, it falls outside the scope of this research.

\section{Operational Domain Model}\label{sec:ODM}

\begin{figure}
\centering
\includegraphics[width=\textwidth]{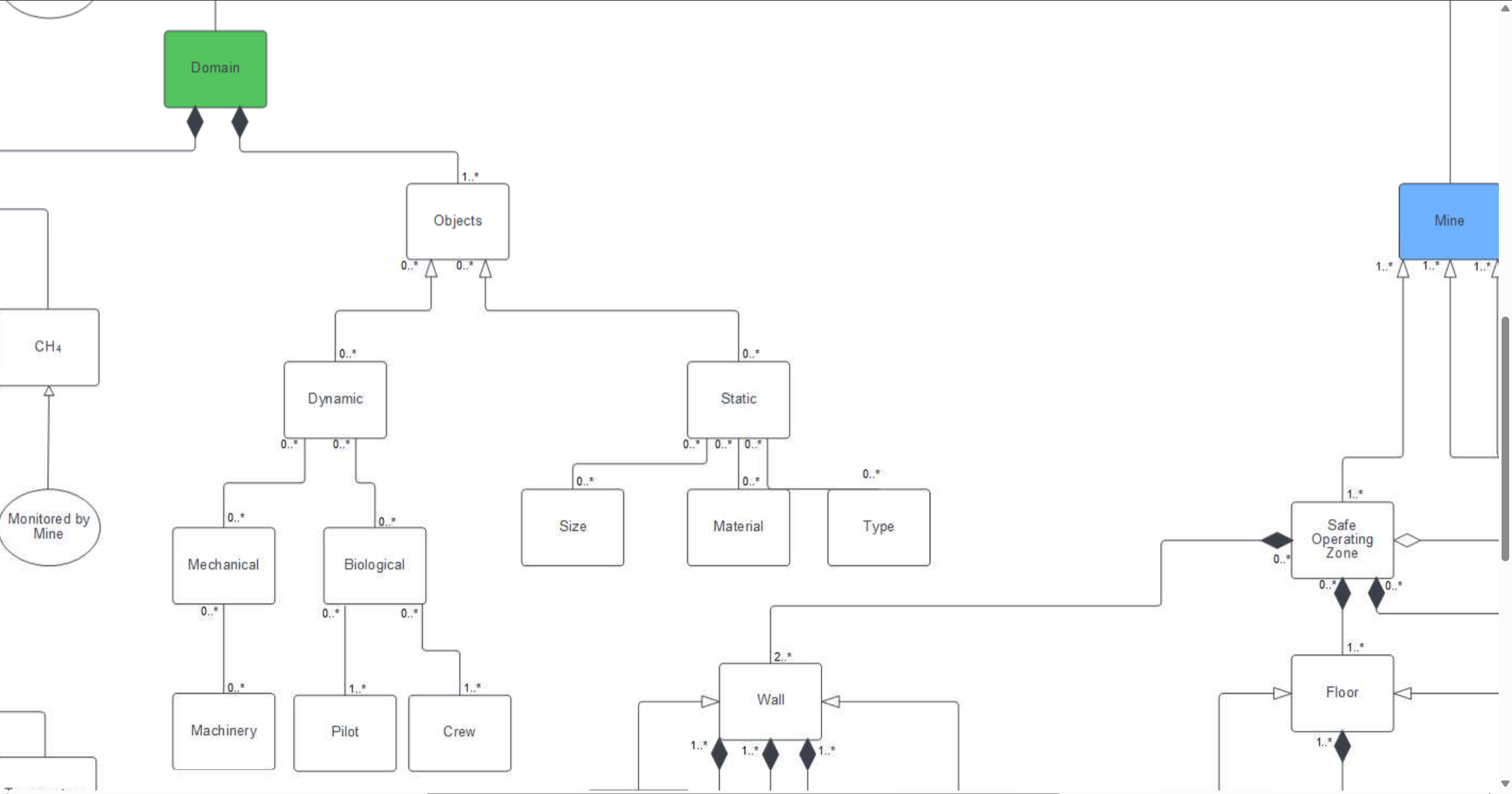}
\caption{Zoomed view of ODM for an underground mine. See \cite{asumi2024} for the full ODM} \label{fig1}
\end{figure}

Fig. \ref{fig1} shows a zoomed view of a portion of an ODM for an underground mine. The ODM offers a structured representation of operational scenarios, environmental conditions, domain-specific factors, and mine structures. The total number of possible test case combinations derived from the ODM is calculated by analysing the key categories and their branches. 

 For example, in the ODM, the \textit{Operational Time} category combines factors such as time metrics, restrictions, and duration metrics. Within the \textit{Environment category}, various factors are considered, including airflow, visibility, moisture, temperature, dust, light, and methane ($CH_4$), each branching into multiple subcategories. The \textit{Operational Domain category} also includes three types of dynamic objects (Machinery, Pilot, and Crew) and static objects. Additionally, the \textit{Mine Structure category} accounts for elements such as ceiling/roof, wall materials, and floor materials, each offering multiple options. The combination of these variables results in thousands of unique test case possibilities. While this systematic approach ensures comprehensive situation coverage, simulating every combination is often infeasible due to cost and resource constraints. Therefore, we first need to select which combinations to test. To do this we begin by creating a situation hyperspace for our SCALOFT testing approach.

\subsection{ODM into Target Situation Hyperspace}

Several challenges arise when justifying the ODM-based approach to defining the target situation hyperspace. First, the situation hyperspace is limitless \cite{Tahir2023}, so there is no perfect or complete solution. This means we must decide which specific areas to focus on for now. In reality, defining a situation hyperspace means choosing certain areas to focus on from an infinite set. While this approach does not provide a clear answer to ``how much is enough?'' it justifies starting with specific priorities since limited resources require us to make informed decisions about where to begin \cite{AAIP2023}.

\begin{figure}
\centering
\includegraphics[width=\textwidth]{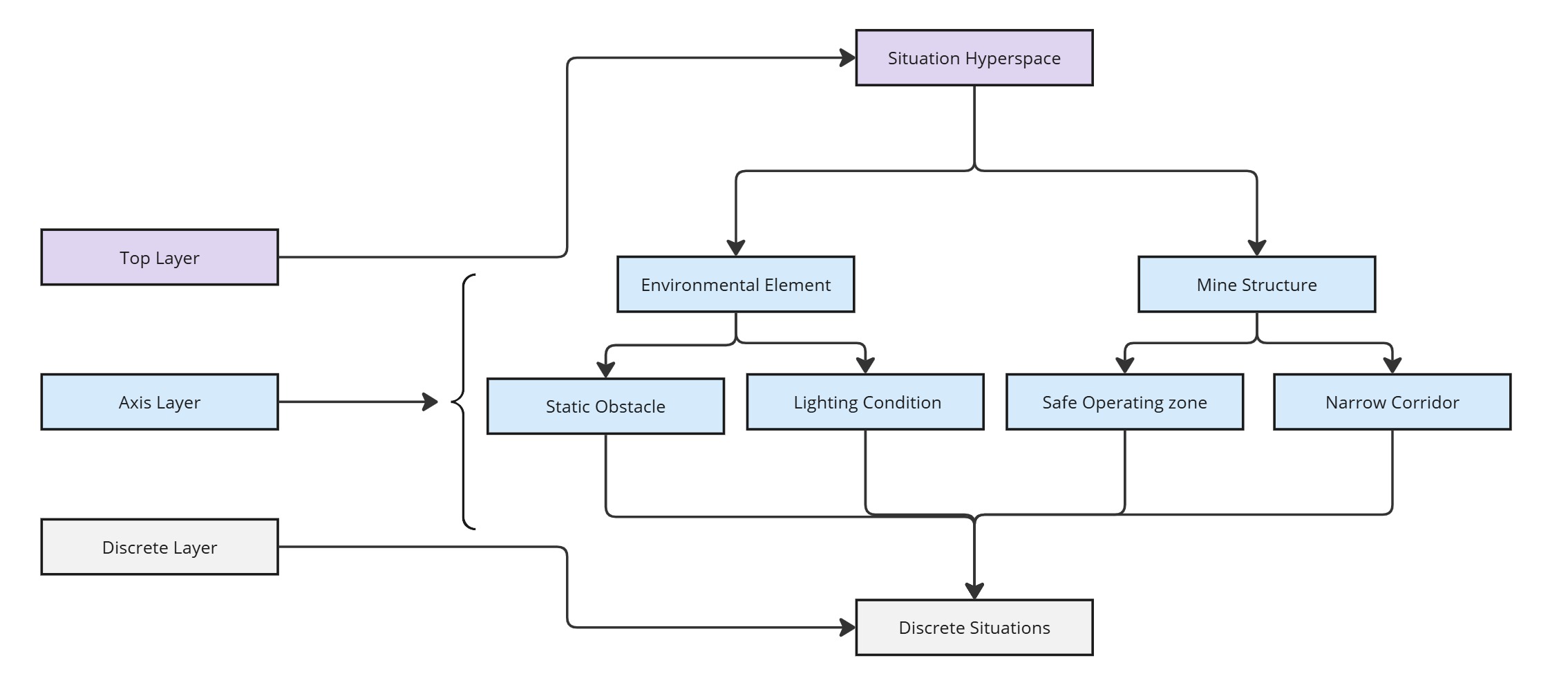}
\caption{Situation hyperspace inspired from \cite{Tahir2023}} \label{fig2}
\end{figure}In our study, the situation hyperspace has been constructed in a methodical way so that our SCALOFT testing approach can systematically navigate through it to generate situations for AAD. In Fig. \ref{fig2}, the top layer of the situation hyperspace shows the main axis, which is split into two parts: the environmental conditions axis and the mine structure axis. Under the Environment category, the emphasis is placed on the light level and static obstacle, representing objects that do not undergo dynamic changes in their position or state. From the mine structure category, the initial coverage target is limited to safe zone i.e. starting position of the AAD and narrow corridors. While more axes could be added, we are currently using only these two as we develop our approach. We are developing the SCALOFT testing system using ROS and Gazebo \cite{osrf_2019_gazebo,ros_2020_rosorg}, and keeping the number of situation elements and their combinations small made it easier to manage and code.

\section{Proposed Testing Approach}

A well-founded safety case template is required to ensure confidence in the safety of autonomous systems (AS). The SACE (Safety Assurance of Autonomous Systems in Complex Environments) guideline offers a structured approach to achieve this \cite{hawkins2022guidance}. It includes a set of safety case templates and a process designed to integrate safety assurance into the system's development while producing evidence to demonstrate that the system operates safely within acceptable limits.

Our proposed testing approach SCALOFT is designed to address the challenge posed in SACE ID-E: Activity 29 — ``Do the test cases sufficiently cover the range of potential operating scenarios for the Autonomous system?''. 

The next section describes our testing environment, its configuration and our proposed testing methodology.

\subsection{ALOFT Setup}

\begin{figure}[h!]
    \centering
    \subfloat[Entrance view with interior layout.]{\includegraphics[width=0.50\textwidth, height=0.25\textheight]{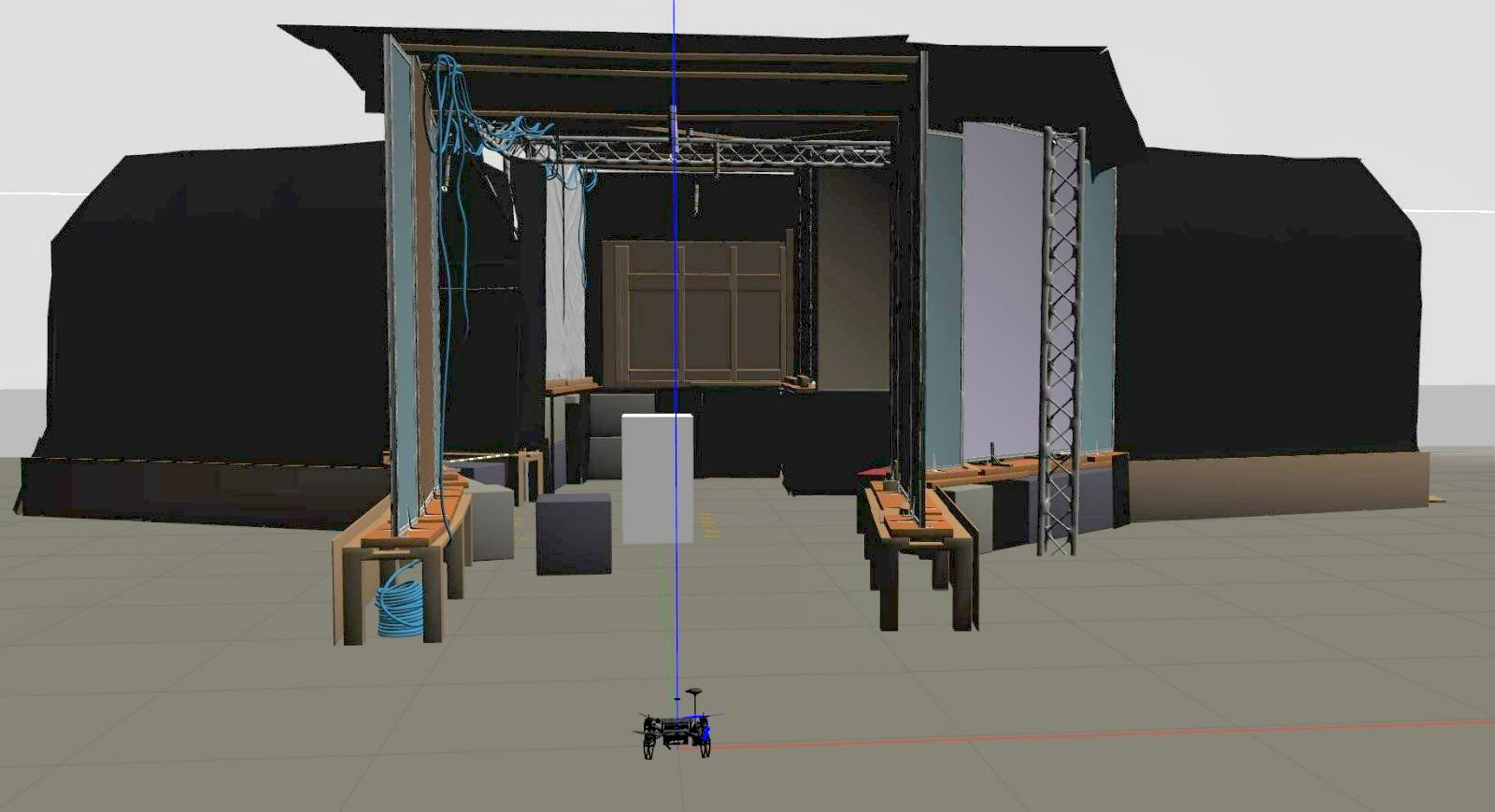} }
    \hfill
    \subfloat[Top-down view with waypoints.]{\includegraphics[width=0.45\textwidth, height=0.25\textheight]{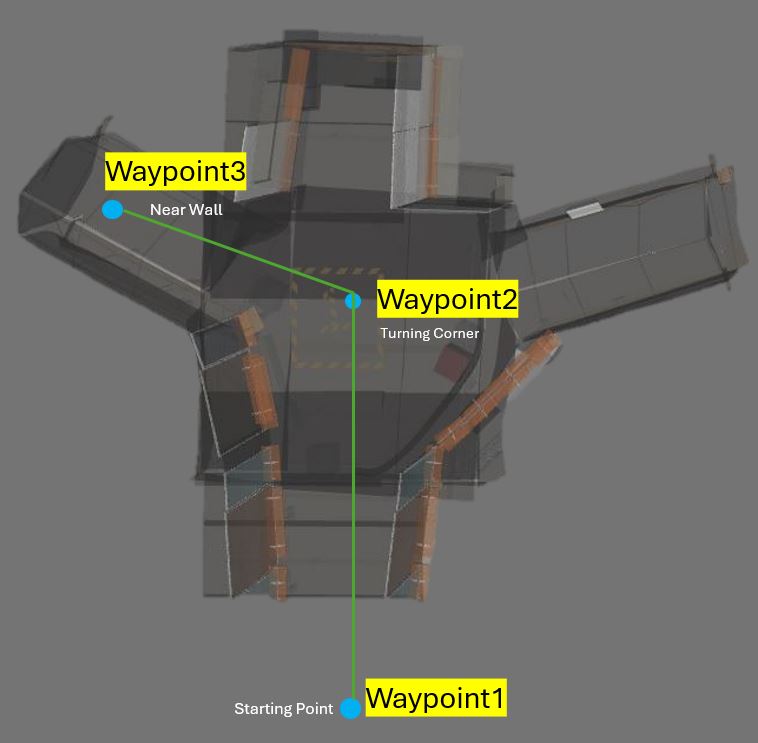} }
    \caption{ALOFT setup}
    \label{fig:ISA}
\end{figure}

Our objective is to use the ALOFT: Self-Adaptive Drone Controller testbed~\cite{imrie2024aloft} for situation coverage-based safety testing of AAD. ALOFT is a digital twin with a full physics engine and 3D simulation. It provides a simulation setup of a mine environment, constructed from 3D laser scans of a mine recreated in a research lab. The environment can be varied in simulation for situation-based testing. ALOFT targets the PX4 Vision V1.5 quadcopter, which uses the PX4-Autopilot flight controller software  \cite{PX4VisionKit2023}. This system allows external flight control via a companion computer, which communicates with the flight control unit and employs ROS for navigation (see fig. \ref{fig:ISA}). ALOFT contains a self-adaptive controller and supports runtime data collection, which can be used for detailed post-flight analysis. The project is available on GitHub at \href{https://github.com/uoy-research/ALOFT}{this repository}. 

In SCALOFT, the AAD is tasked with surveying inside the mine. The mission begins in an open space designated as the safe operating zone with or without a human present at the entrance in Fig. \ref{fig:ISA}(b). The drone then navigates into the mine marked as a green line, turning a corner and flying near a wall while conducting the survey. Under ideal conditions, it follows the same trajectory back to the safe operating zone, ensuring a controlled and safe return.

\subsection{SCALOFT}
\begin{figure}[h!]
\centering
\includegraphics[width=10cm]{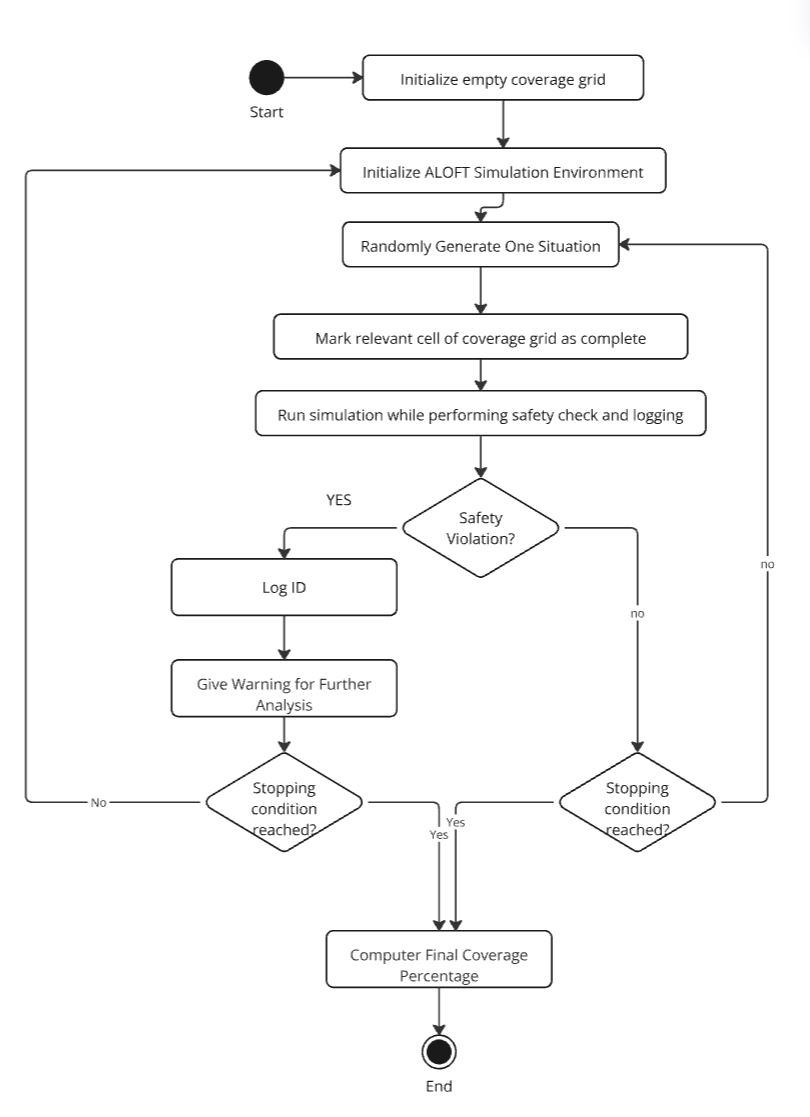}
\caption{Our proposed testing approach} \label{fig:SCALOFT}
\end{figure} Fig \ref{fig:SCALOFT} illustrates the workflow of SCALOFT testing approach. It begins with initializing an empty coverage grid and the ALOFT simulation environment. In our study, an initial version of the situation coverage grid can be created using five axes, each with two possible values (see table \ref{tab2}). 

 Combining these factors results in $2^5 = 32$ discrete situations (see table \ref{tab3}), providing a structured way to explore different test cases. The system then iteratively generates random situations, marks the corresponding cells in the coverage grid, and simulates the drone's behaviour while performing safety checks. In this case, the initial safety requirements (SRs) are defined at a high level, as situation coverage-based safety testing does not necessitate detailed requirement specifications. Instead, generic safety requirements are sufficient to guide the testing process. The safety requirements considered are as follows:
 \begin{itemize}
    \item \textbf{SR1}: The drone shall avoid collisions under all operating conditions.
    \item \textbf{SR2}: Upon detecting a person within a specified distance, the drone shall reduce its speed and avoid collision.
\end{itemize}

During safety checking, any detected safety violation is logged for further analysis, and a warning is issued. The process continues until a predefined stopping condition is reached, after which the final coverage percentage is calculated as the ratio of tested situations to the total generated situations (see fig. \ref{fig:SCALOFT}).

\begin{table}[h!]
\renewcommand{\arraystretch}{1.25}
\centering
\caption{Situation coverage grid}\label{tab2}
\setlength{\tabcolsep}{2.5pt} 
\begin{tabular}{p{3.5cm}p{4cm}p{4cm}}
\hline
\textbf{Axis} & \textbf{Value 1} & \textbf{Value 2} \\
\hline
Turning a corner & Mission does not require turning a corner & Mission requires turning a corner \\
Obstacle on path & No & Yes \\
Waypoint placement & All waypoints in open space & At least one waypoint near a wall \\
Lighting condition & Default & Total darkness \\
Human presence & Present & Absent  \\
\hline
\end{tabular}
\end{table}

\begin{table}[h!]
\centering
\caption{Discrete situations}\label{tab3}

\renewcommand{\arraystretch}{1.5} 
\begin{tabular}{|p{1cm}|p{2cm}|p{2cm}|p{2cm}|p{2cm}|p{2cm}|}
\hline
\textbf{ID} & \textbf{Turning } & \textbf{Obstacle} & \textbf{Waypoint Placement} & \textbf{Lighting Condition} & \textbf{Human Presence} \\
\hline
1 & No & No & Open space & Default & Yes \\
\hline
2 & No & No & Open space &     Dark & No \\
\hline
3 & No & No &  Near a wall & Default & No \\
\hline
\multicolumn{6}{|c|}{\dots (Situations 4 to 31)} \\
\hline
32 & Yes & Yes & Near a wall & Dark & No \\
\hline
\end{tabular}
\end{table}

\begin{figure}[h!]
    \centering
    \subfloat[Ideal condition(Drone flies a height of 1.65 m unless there is an obstcale in it's path)]{\includegraphics[width=0.45\textwidth, height=0.25\textheight]{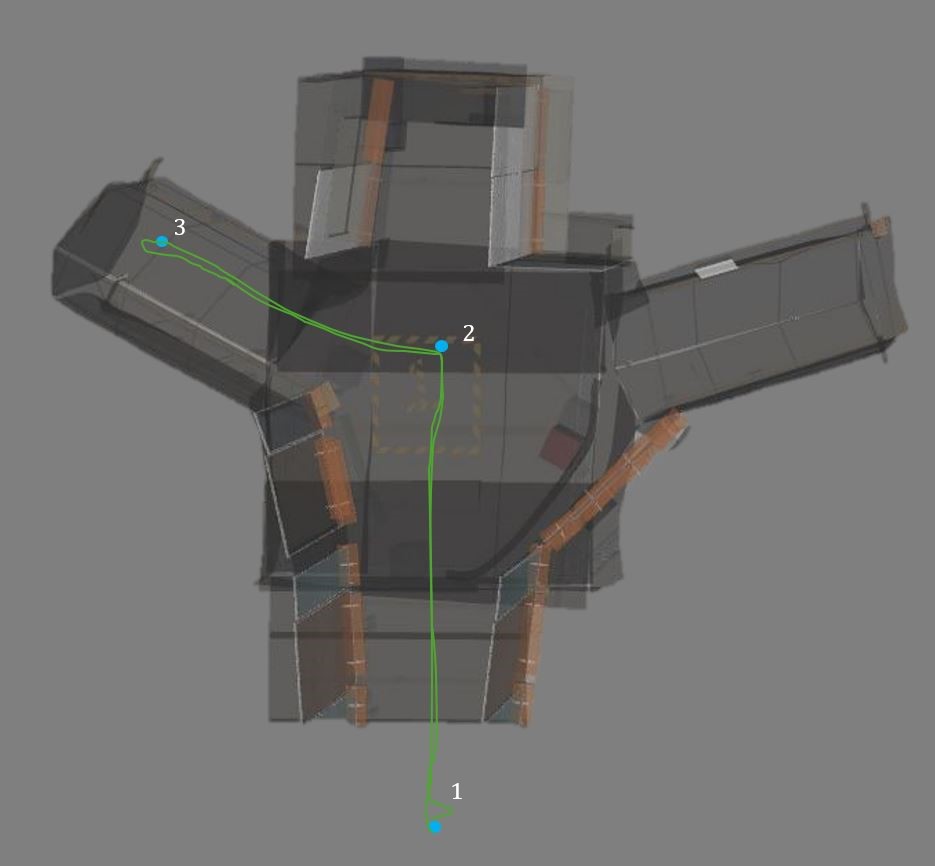} }
    \hfill
    \subfloat[Person at the entrance and an Obstacle (0.5m× 0.5m × 1.8m) on obvious path ]{\includegraphics[width=0.45\textwidth, height=0.25\textheight]{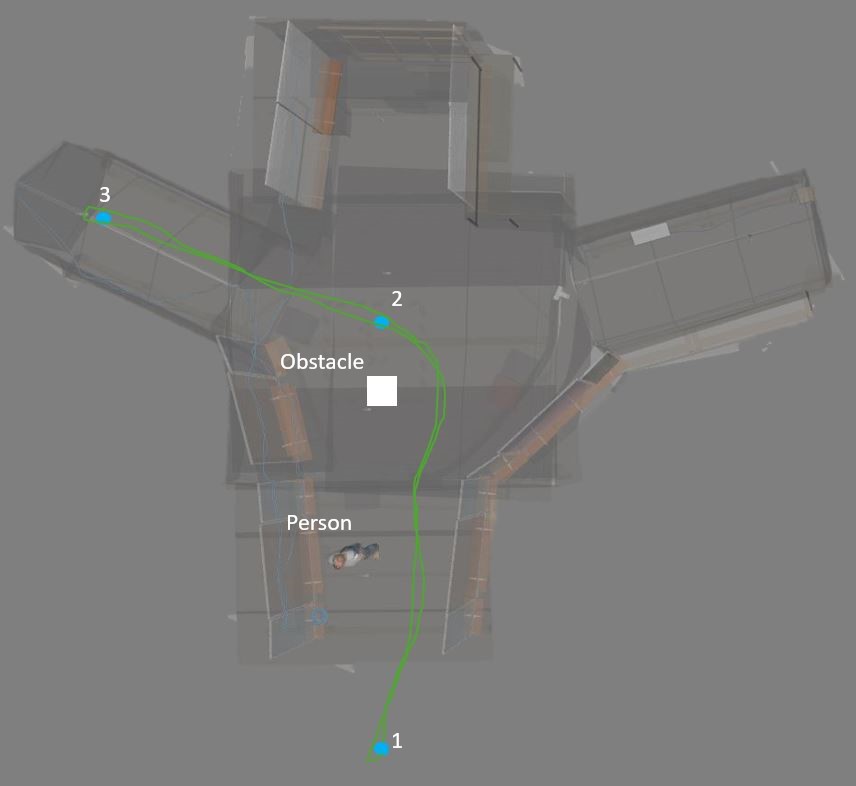} }
    \vfill
    \subfloat[An obstacle (0.5m× 0.5m × 1.8m) positioned on obvious path]{\includegraphics[width=0.45\textwidth, height=0.25\textheight]{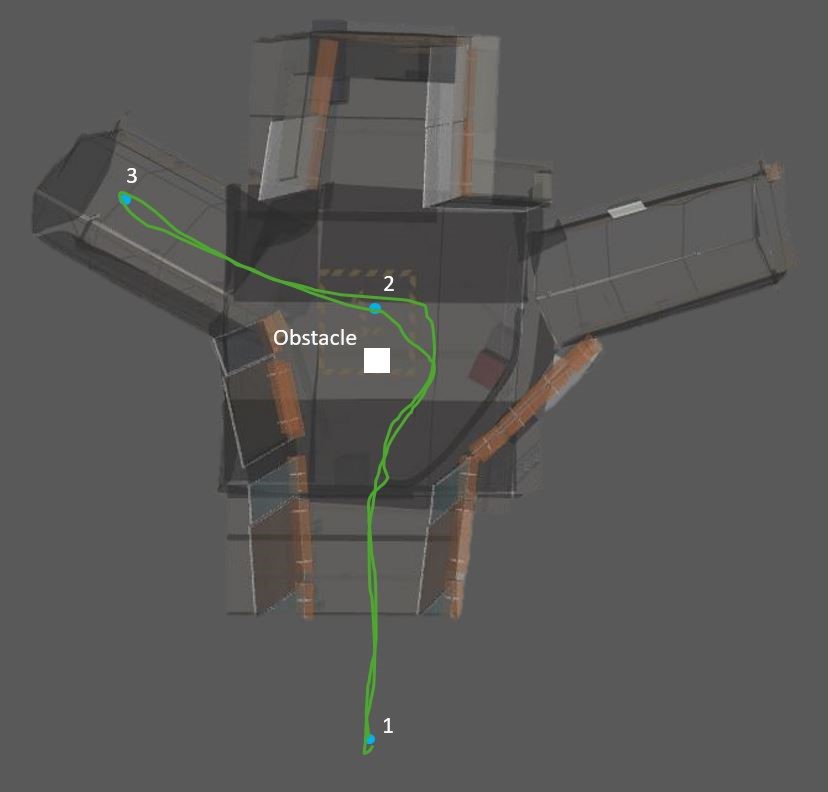} }
    \hfill
    \subfloat[Only one person positioned at the entrance]{\includegraphics[width=0.45\textwidth, height=0.25\textheight]{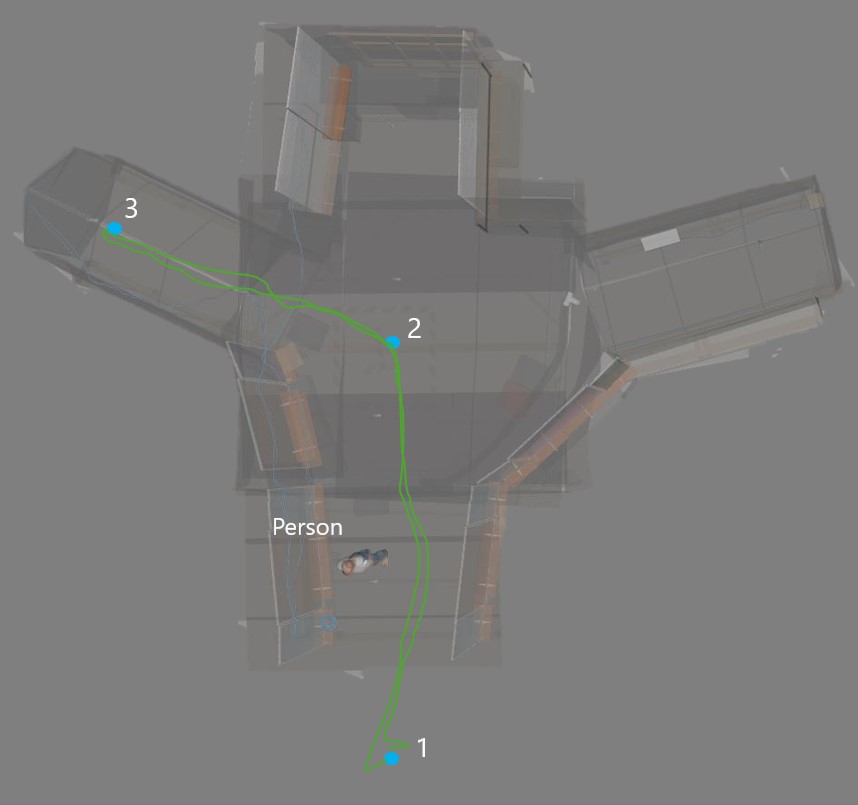} }
    \caption{Comparison of drone's journey in different conditions}\label{multifig}
\end{figure}

Fig. \ref{multifig}(a) shows the drone's trajectory under normal conditions — with no obstacles in its path and no human present. The AAD successfully completed its mission from the safe zone (waypoint 1) through waypoints 2 and 3, back to 2 and back home to waypoint 1 (marked as blue dots). During its mission, the drone only had knowledge of its next waypoint at any given time. In fig. \ref{multifig} (b),(c) and (d), the SCALOFT testing approach monitors the drone’s journey through different test cases in default light condition, recording its path and interactions in the simulation. Here, default lighting conditions refer to an artificial overhead light that illuminates the entire mine. The green lines in fig. \ref{multifig} represent the actual recorded positions of the drone during its journey inside the ALOFT environment.

\begin{figure}
\centering
\includegraphics[width=0.5\textwidth,height=4cm]{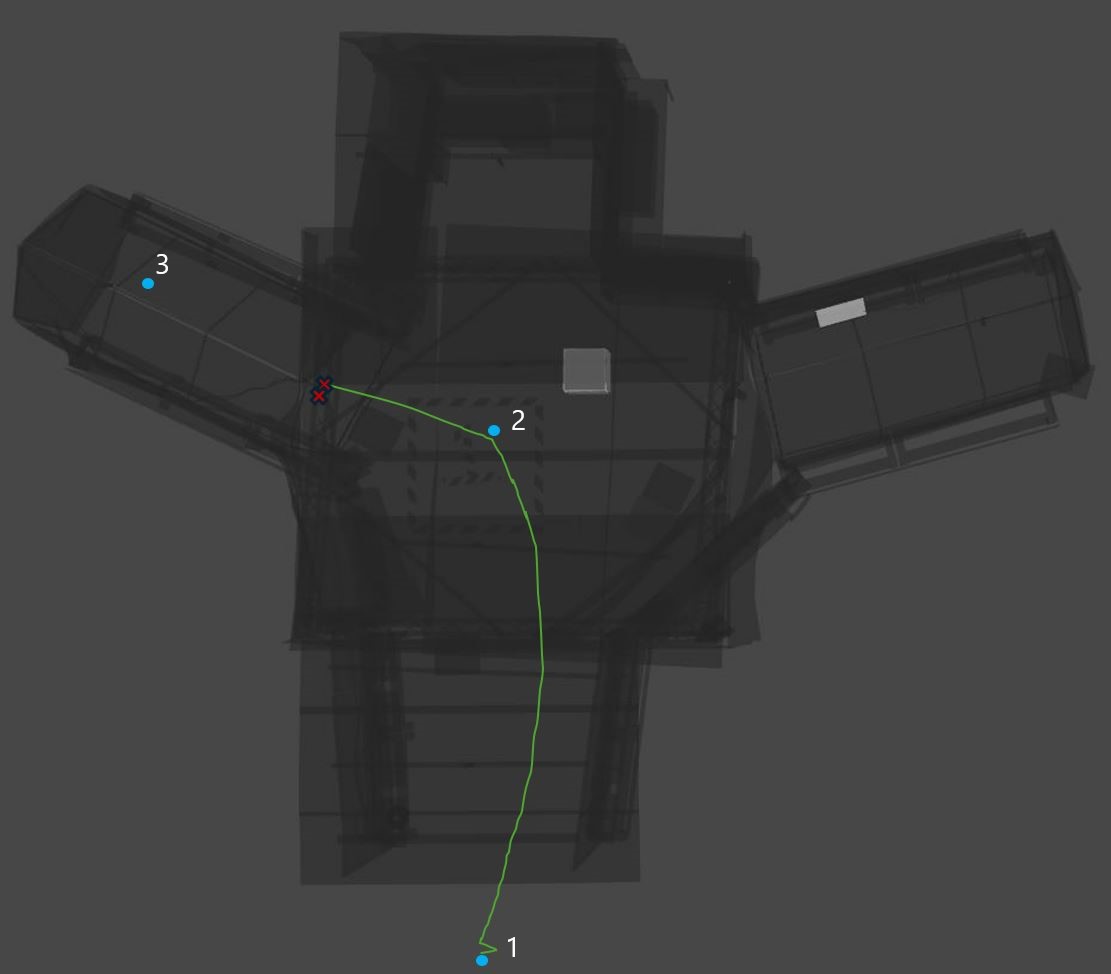}
\caption{Drone's journey in no light setting} \label{no light}
\end{figure}

We also tested these conditions in a completely dark environment, as shown in Fig. \ref{no light}. Even in total darkness, the drone was able to follow the waypoint for a short distance and continue flying while turning a corner. However, it eventually collided with a bar inside the mine marked in red in  fig. \ref{no light}. Since the drone is equipped with a depth camera, it was able to detect its surroundings to some extent, which is why it was able to fly for a brief period \cite{intel_realsense_2025}. Information about this specific test cases, along with all other situations with a unique ID representing collision position and time, was recorded in a JSON file for further analysis. The log file also contains data on the total possible test cases and the total test cases generated during the drone's journey, which can be used to calculate the coverage percentage at the end. The complete log file, along with implementation details and results, is accessible via the associated \href{https://github.com/uoy-research/SCALOFT}{ GitHub repository}.

\section{Performance Evaluation}

SCALOFT addresses dynamic safety assurance in challenging scenarios by collecting data on safety violations using continuous runtime monitoring \cite{tamura2013towards}. While it has shown promise in handling initial coverage targets, its scalability and adaptability to more complex situational hyperspaces requires further evaluation. In particular, how to select seeded faults for testing, to extend table \ref{tab2}.

Previous studies on situation coverage-based safety testing for AVs (e.g. \cite{Tahir2022,nawshin2023}) have often relied on unsystematic or ad hoc methods for selecting seeded faults during performance evaluation. In contrast, we introduce a more structured approach by using HAZOP (HAZard OPerability analysis) guidewords \cite{standard2001hazard} to systematically derive meaningful fault injection scenarios within the ALOFT simulation environment (see table \ref{tab:fault_analysis}). Deviations were systematically identified using HAZOP guidewords such as LATE, UNINTENDED, and MORE. These guidewords serve as triggers for identifying potential failure modes and initiating investigations on how faults may propagate through the system. By combining each guideword with relevant system parameters (such as introducing a timing delay, see table \ref{tab:fault_analysis}), we created a Deviation Matrix that shows how the system could behave differently from what was expected, helping us uncover possible faults. The observed safety violations across all fault injection scenarios indicate that the seeded faults were successfully triggered and validate the effectiveness of the SCALOFT testing approach for small faults.
\begin{table}[h!]
    \centering
    \fontsize{8pt}{8pt}\selectfont
    \renewcommand{\arraystretch}{2} 
    \caption{Performance evaluation of SCALOFT}
    \setlength{\tabcolsep}{6pt} 

    \begin{tabular}{p{2cm}p{3cm}p{4cm}p{1.4cm}}
\hline
\textbf{HAZOP Guidewords} & \textbf{Injected Faults}  & \textbf{Outcome} & \textbf{Safety Violation} \\
\hline
LATE: Relative to the clock time & Delay detecting human by 3 seconds  & Drone failed to reduce speed due to late detection  & SR2 \\

UNINTENDED: Unintended activation & Simulate false collision every 20 sec   & After experiencing a false collision detection, the drone actually collided with a wall & SR1 \\

MORE: Quantitative increase & Increase goal threshold  & An increased goal threshold leading the drone to navigate incorrectly and collide before reaching the next waypoint & SR1 \\
\hline
\end{tabular}
\label{tab:fault_analysis}
\end{table}
        
\section{Conclusion}

SCALOFT represents an initial yet promising approach to situation coverage-based safety analysis for autonomous aerial drones in mine environments. By leveraging structured test case generation and real-time safety monitoring, it effectively identifies safety violations and ensures a systematic evaluation of drone behaviour in dynamic scenarios. The initial performance evaluation results with seeded faults highlight the system's ability to detect small faults. We then introduced a more structured approach to systematically derive meaningful fault injection scenarios (seeded faults) by using HAZOP guidewords. This systematically introduces faults at the system behaviour level such as timing delays or false positive detections. It allows us to run the fault in simulation and analyse the outcomes increasing the situation-coverage level achieved.

However, as the situation space expands, several challenges must be addressed. The complexity of real-world scenarios necessitates a more scalable approach to defining and testing situation hyperspaces. The current model, while structured, is limited in its ability to handle an infinite number of potential operational conditions. 
\\\\
\noindent\textbf{Acknowledgements}: 
This work was supported by the Centre for Assuring Autonomy, a partnership between Lloyd’s Register Foundation and the University of York (https://www.york.ac.uk/assuring-autonomy/)

%
%

\bibliographystyle{splncs04}
\bibliography{references}

\begin{thebibliography}{10}
\providecommand{\url}[1]{\texttt{#1}}
\providecommand{\urlprefix}{URL }
\providecommand{\doi}[1]{https://doi.org/#1}

\bibitem{PX4VisionKit2023}
Px4 vision kit (2023), available at: \url{https://docs.px4.io/main/en/complete_vehicles/px4_vision_kit.html[Accessed: 2025-01-28]}

\bibitem{intel_realsense_2025}
Range of d435 depth camera in darkness (2025), \url{https://support.intelrealsense.com/hc/en-us/community/posts/360037389393-range-of-D435-depth-camera-in-darkness-or-poor-lighting-condition}, accessed: 2025-03-20

\bibitem{abdessalem2018testing}
Abdessalem, R.B., Nejati, S., Briand, L.C., Stifter, T.: Testing vision-based control systems using learnable evolutionary algorithms. In: Proceedings of the 40th International Conference on Software Engineering. pp. 1016--1026 (2018)

\bibitem{AAIP2023}
Alexander, R.: {AAIP Robot Demonstrator Project Testing Strategy Report} (2023), unpublished

\bibitem{Rob2015}
Alexander, R., Hawkins, H.R., Rae, A.J.: Situation coverage--a coverage criterion for testing autonomous robots. Tech. rep., Department of Computer Science, University of York (2015)

\bibitem{babikian2020}
Babikian, A.A.: Automated generation of test scenario models for the system-level safety assurance of autonomous vehicles. In: Procs of 23rd ACM/IEEE Conference on Model Driven Engineering Languages and Systems (2020)

\bibitem{Hawkins2019}
Hawkins, H., Alexander, R.: Situation coverage testing for a simulated autonomous car--an initial case study. arXiv preprint arXiv:1911.06501  (2019)

\bibitem{hawkins2022guidance}
Hawkins, R., Osborne, M., Parsons, M., Nicholson, M., McDermid, J., Habli, I.: Guidance on the safety assurance of autonomous systems in complex environments (sace). arXiv preprint arXiv:2208.00853  (2022)

\bibitem{asumi2024}
Hodge, V.J.: {Assuring the Safety of UAVs for Mine Inspection (ASUMI)}, {ODM available at \url{https://www-users.york.ac.uk/~vjh5/myPapers/ASUMI_ODM.pdf}}

\bibitem{hodge2021deep}
Hodge, V.J., Hawkins, R., Alexander, R.: Deep reinforcement learning for drone navigation using sensor data. Neural Computing and Applications,  \textbf{33}(6),  2015--2033 (2021)

\bibitem{huang2017safety}
Huang, X., Kwiatkowska, M., Wang, S., Wu, M.: Safety verification of deep neural networks. In: Computer Aided Verification: 29th International Conference, CAV 2017, Heidelberg, Germany, July 24-28, 2017. pp. 3--29. Springer (2017)

\bibitem{imrie2024aloft}
Imrie, C., Howard, R., Thuremella, D., Proma, N.M., Pandey, T., Lewinska, P., Cannizzaro, R., Hawkins, R., Paterson, C., Kunze, L., et~al.: Aloft: Self-adaptive drone controller testbed. In: SEAMS'24: Proceedings of the 19th Symposium on Software Engineering for Adaptive and Self-Managing Systems. ACM (2024)

\bibitem{iqbal2015environment}
Iqbal, M.Z., Arcuri, A., Briand, L.: Environment modeling and simulation for automated testing of soft real-time embedded software. Software \& Systems Modeling  \textbf{14},  483--524 (2015)

\bibitem{katz2017reluplex}
Katz, G., Barrett, C., et~al.: Reluplex: An efficient smt solver for verifying deep neural networks. In: Procs of 29th International Conference on Computer Aided Verification, CAV 2017, Heidelberg, Germany, July 24-28. pp. 97--117 (2017)

\bibitem{kurakin2018adversarial}
Kurakin, A., Goodfellow, I.J., Bengio, S.: Adversarial examples in the physical world. In: Artificial intelligence safety and security, pp. 99--112. Chapman and Hall/CRC (2018)

\bibitem{majzik2019towards}
Majzik, I., Semer{\'a}th, O., Hajdu, C., et~al.: Towards system-level testing with coverage guarantees for autonomous vehicles. In: 2019 ACM/IEEE 22nd International Conference on Model Driven Engineering Languages and Systems (MODELS). pp. 89--94. IEEE (2019)

\bibitem{micskei2012concept}
Micskei, Z., Szatm{\'a}ri, Z., Ol{\'a}h, J., Majzik, I.: A concept for testing robustness and safety of the context-aware behaviour of autonomous systems. In: Procs of 6th KES International Conference: Agent and Multi-Agent Systems, Technologies and Applications (KES-AMSTA), June 25-27. pp. 504--513. Springer (2012)

\bibitem{nguyen2012evolutionary}
Nguyen, C.D., Miles, S., et~al.: Evolutionary testing of autonomous software agents. Autonomous Agents and Multi-Agent Systems  \textbf{25},  260--283 (2012)

\bibitem{osrf_2019_gazebo}
OSRF: Gazebo. \url{http://gazebosim.org/} (2019)

\bibitem{pei2017deepxplore}
Pei, K., Cao, Y., Yang, J., Jana, S.: Deepxplore: Automated whitebox testing of deep learning systems. In: proceedings of the 26th Symposium on Operating Systems Principles. pp. 1--18 (2017)

\bibitem{nawshin2023}
Proma, N.M., Alexander, R.: Systematic situation coverage versus random situation coverage for safety testing in an autonomous car simulation. In: Procs of the 12th Latin-American Symposium on Dependable and Secure Computing. p. 208–213. LADC '23 (2023), \url{https://doi.org/10.1145/3615366.3625077}

\bibitem{proma2024situation}
Proma, N.M., Hodge, V.J., Alexander, R.: Situation coverage based safety analysis of an autonomous aerial drone in a mine environment. In: The Yorkshire Innovation in Science and Engineering Conference (YISEC) 2024. York (2024)

\bibitem{ros_2020_rosorg}
ROS: Ros.org | powering the world's robots. \url{https://www.ros.org/} (2020)

\bibitem{ryan2024safety}
Ryan, P., Badyal, A., et~al.: Safety assurance challenges for autonomous drones in underground mining environments. In: Towards Autonomous Robotic Systems: 25th Annual Conference, TAROS 2024, London, UK, August 21–23, Proceedings. pp. 169--181. Springer (2025). \doi{10.1007/978-3-031-72059-8_15}

\bibitem{shakhatreh2019unmanned}
Shakhatreh, H., Sawalmeh, A.H., et~al.: Unmanned aerial vehicles (uavs): A survey on civil applications and key research challenges. Ieee Access  \textbf{7},  48572--48634 (2019)

\bibitem{standard2001hazard}
Standard, B., IEC61882, B.: Hazard and operability studies (hazop studies)-application guide. International Electrotechnical Commission  (2001)

\bibitem{Tahir2022}
Tahir, Z., Alexander, R.: Intersection focused situation coverage-based verification and validation framework for autonomous vehicles implemented in carla. In: Procs Modelling and Simulation for Autonomous Systems: 8th International Conference, MESAS 2021, October 13--14, 2021. pp. 191--212 (2022)

\bibitem{Tahir2023}
Tahir, Z.: Situation hyperspace — using a simulated world to obtain situation coverage for av safety assurance (2023), available at: https://assuringautonomy.medium.com/situation-hyperspace-using-a-simulated-world-to-obtain-situation-coverage-for-av-safety-assurance-39fa5ea203cd [Accessed: 2025-01-28]

\bibitem{tamura2013towards}
Tamura, G., Villegas, N.M., M{\"u}ller, H.A., et~al.: Towards practical runtime verification and validation of self-adaptive software systems. In: Software Engineering for Self-Adaptive Systems II: International Seminar, Dagstuhl Castle, Germany, October 24-29, 2010. pp. 108--132. Springer (2013)

\bibitem{tian2018deeptest}
Tian, Y., Pei, K., Jana, S., Ray, B.: Deeptest: Automated testing of deep-neural-network-driven autonomous cars. In: Proceedings of the 40th international conference on software engineering. pp. 303--314 (2018)

\bibitem{ulbrich2015defining}
Ulbrich, S., Menzel, T., et~al.: Defining and substantiating the terms scene, situation, and scenario for automated driving. In: 2015 IEEE 18th international conference on intelligent transportation systems. pp. 982--988. IEEE (2015)

\end{thebibliography}

\end{document}